# NaroNet: Discovery of novel tumor microenvironment elements from highly multiplexed images.

Daniel Jiménez-Sánchez[1,2], Mikel Ariz[1], Hang Chang[2], Xavier Matias-Guiu[3,4], Carlos E. de Andrea[5] & Carlos Ortiz-de-Solórzano[1,*]

[1]*Solid Tumors and Biomarkers Program, IDISNA, and Ciberonc, Center for Applied Medical Research, University of Navarra, 31008, Pamplona, Spain*

[2]*Biological Systems and Engineering Division, Lawrence Berkeley National Laboratory, CA, 94720, Berkeley, USA*

[3]*Department of Pathology, Hospital Universitari Arnau de Vilanova, University of Lleida, IRB-Lleida, Lleida, 25198, Spain*

[4]*Department of Pathology, Hospital Universitari de Bellvitge, IDIBELL, CIBERONC, 08908, Barcelona, Spain*

[5]*Department of Pathology, IDISNA, Ciberonc, Clínica Universidad de Navarra, University of Navarra, 31008, Pamplona, Spain*

[*]*Corresponding author*

**Abstract**

Many efforts have been made to discover tumor-specific microenvironment elements (TMEs) from immunostained tissue sections. However, the identification of yet unknown but relevant TMEs from multiplex immunostained tissues remains a challenge, due to the number of markers involved (tens) and the complexity of their spatial interactions. We present NaroNet, which uses machine learning to identify and annotate known as well as novel TMEs from self-supervised embeddings of cells, organized at different levels (local cell phenotypes and cellular neighborhoods). Then it uses the abundance of TMEs to classify patients based on biological or clinical features. We validate NaroNet using synthetic patient cohorts with adjustable incidence of different TMEs and two cancer patient datasets. In both synthetic and real datasets, NaroNet unsupervisedly identifies novel TMEs, relevant for the user-defined classification task. As NaroNet requires only patient-level information, it renders state-of-the-art computational methods accessible to a broad audience, accelerating the discovery of biomarker signatures.

**Introduction**

Computational Pathology (CP) is a thriving field aimed at automating the *in situ* analysis of tumors. Two main CP strategies exist, weakly-supervised Deep Learning (WSDL) and Single-Cell Analysis (SCA). WSDL methods focus on the analysis of "architectural" tissue patterns in standard Hematoxylin & Eosin (H&E) stained tissues. To this end, early WSDL methods trained Deep Learning (DL) segmentation models[1,2,3] to identify tumor components from extensive pixel-level pathologist annotations[4]. More recently, advances in digital pathology and whole slide imaging (WSI) have facilitated the compilation of large image datasets, allowing a shift in WSDL towards the use of DL not to classify tissue structures, but to *blindly* extract prominent tumor features from the raw H&E stained tissue data, and use those features to classify patients based on patient-level, clinical labels[5]. These weakly-supervised DL models show unprecedented predictive ability, often outperforming human experts[6,7,8]. Nowadays, efforts are on their way to make WSDL interpretable, i.e. able to unsupervisedly identify which tissue elements are more relevant in classification[9].

Single-Cell Analysis (SCA) emerged in the context of the research for novel cancer biomarkers. This is a laborious task that requires selecting potential targets from *in silico* data, validating these targets *in situ*, and statistically confirming that they are reliably related to a specific biological effect[10]. Traditionally, biomarkers have been validated individually. Nowadays, novel highly multiplexed tissue processing and imaging technologies, such as imaging mass cytometry (IMC) or multiplex immunofluorescence (MI), allow simultaneous validation of high number (>20) of biomarkers[11,12,13]. These complex biomarker *signatures* provide a deeper understanding of the tumor microenvironment, which could in turn aid to understand the biology of the tumor and develop new therapies. However, the high number of markers involved and the complexity of their interactions prevents the use of classical manual tools[14] that must be replaced by automated systems able to process vast amounts of data, learning complex cell interaction patterns easily hidden to the expert's eye, and associating those interactions with the patients' diagnosis or prognosis[15,16]. HistoCat[17,18] was the first method that applied SCA to study the tumor microenvironment from highly multiplexed image data. To this end, HistoCat segments all the cells in the tissue and extracts their phenotype (marker expression and morphology), and the number and phenotype of the neighboring cells. This information is then used to cluster cells with similar phenotypes, similar to what is done in flow cytometry[19]. This way a tissue/patient is represented by a vector that contains the abundance of each cluster type. Schürch et al.[20] went one step further by introducing the concept of 'neighborhoods', or higher-order interactions between one or more cellular phenotypes. The authors build topological networks containing these interactions and apply a graph-based clustering approach[21] to assign groups of cells to different neighborhoods. Once several tissues/patients are characterized by integrating the information of cell-phenotypes and neighborhood interactions, they group patients with similar characteristics. These SCA methods use the cell as the basic unit of tissue representation, providing a high level of interpretability. However, SCA is a sequential, not learning-based approach. It requires cell segmentation[22,23] and unsupervised clustering[19], to group cells into cell phenotypes and cellular neighborhoods using thresholds chosen *a posteriori* to minimize differences between clusters. Therefore, if images are affected by technical non-linear variabilities caused by the presence of autofluorescence, spectral leakage between biomarkers, low expression of some antigens, etc.[24,20] fixed thresholds are unable to adapt to such differences, complicating the task of biomarker *signature* discovery when the datasets are of heterogeneous origin.

We present NaroNet, an integrated end-to-end trainable, interpretable machine learning (ML) model that automatically identifies local phenotypes, cellular neighborhoods, and interactions between neighborhoods (i.e., tumor microenvironment elements, or TMEs) from highly multiplexed cancer tissue images, and uses the tumor microenvironment information in a classification task of discovery, diagnostic or predictive nature. NaroNet combines features of SCA (use of graphs, interpretable quantification of the tumor microenvironment) and WSDL (patch-based representation, end-to-end learning), addresses some of their limitations, and applies some methodological state-of-the-art ML methods (contrastive learning, graph neural networks, etc.) to learn *in situ* the tumor microenvironment from highly multiplexed tissue images. NaroNet can be used in "discovery mode" to research new biomarker *signatures* of the tumor biology, or to answer clinically relevant questions, e.g. which TMEs are more predictive of the outcome of the patient. Furthermore, using validated biomarker *signatures*, NaroNet can be trained to provide clinicians with interpretable predictions. As NaroNet learns to annotate and identify TMEs that has never seen, the model is inherently interpretable. This is true both at cohort and at individual patient levels, since the TMEs that globally contribute to the classification can be mapped back onto the original images of each patient.

The concept and main elements of NaroNet are illustrated in **Fig. 1** and explained in detail in the **Online methods** section. To validate NaroNet, we first analyzed synthetic sets of multiplex images simulating situations that can be found in real samples (changes in the frequency and/or intensity of a cell marker; affinity or incompatibility between cell phenotypes, etc.). Then we applied our model to learn relevant TMEs and predict molecular and clinically relevant parameters from two real datasets: 336 7-plex images from 12 patients with high-grade endometrial cancer and a publicly available imaging mass cytometry dataset[24] consisting in 281 breast cancer patients.

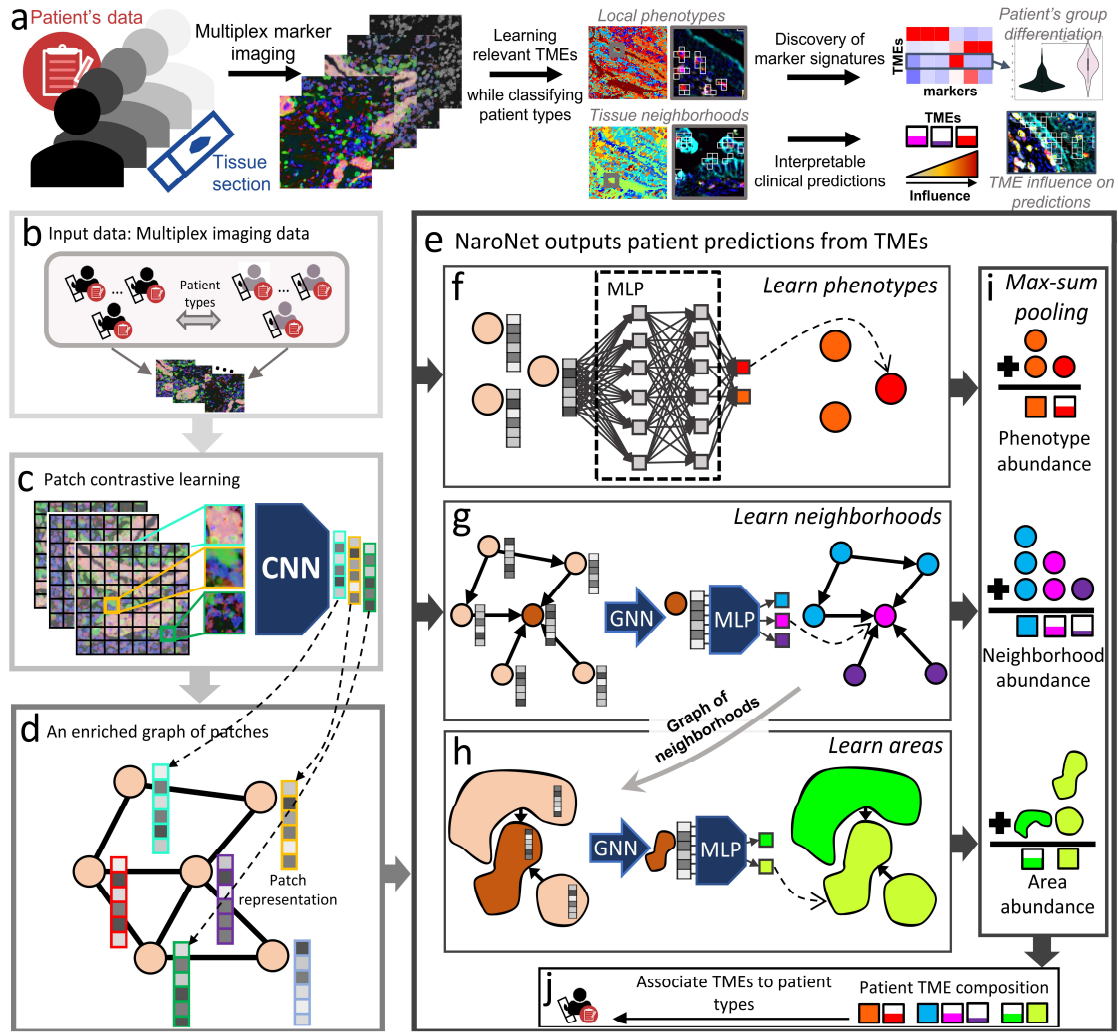

**Fig. 1: Objective-based learning of relevant cancer tissue phenotypes, neighborhoods, and areas from highly multiplexed histological imaging. a.** Scheme of NaroNet's learning and discovery protocol. **b.** The input data consists of multiplex cancer tissue images with associated clinical and pathological information. **c.** The patch contrastive learning module divides images into patches and embeds each patch in a 256-dimensional vector using a CNN unsupervisedly trained to assign similar vectors to patches containing similar biological structures. **d.** An enriched graph of patches is generated that contains the spatial interactions between tissue patches. **e.** The graph of patches is fed to NaroNet: an interpretable ensemble of neural networks that learns phenotypes (**f**), phenotype neighborhoods (**g**), and areas of interaction between neighborhoods (**h**) to classify patients (**j**) based on the abundance of those tumor microenvironment elements (**i**). Legend. CNN: convolutional neural network; MLP: multilayer perceptron; GNN: graph neural network.

# Results

**Synthetic experiments**

Synthetic experiments were performed to evaluate NaroNet's predictive accuracy and interpretability having a quantifiable ground truth. To this end, we used Synplex[25], an in-house developed multiplex immunostained tissue simulator, to create patient cohorts representing different disease paradigms. For each paradigm, a patient cohort of $M = 40 \times O$ patients was generated, being $O$ the number of groups, each one defined by the abundance or interactions between specific TMEs. Each patient was represented by a 800x800 multiplex image that contained 8 cell phenotypes (Ph1-Ph8), defined by the (tunable) probabilistic level of expression of 6 fluorescently labeled markers (Mk1-Mk6) (Supplementary Fig. 1a), size (Supplementary Fig. 1b), and shape (Supplementary Fig. 1c). Four types of cell neighborhoods (Nb1-Nb4) were also defined based on the (adjustable) probabilistic abundance of the six cell phenotypes (Supplementary Fig. 1d), and the (adjustable) interactions between them (Supplementary Fig. 1e). Each neighborhood had a predefined prevalence in the tissue (Supplementary Fig. 1f) and could interact with other neighborhoods (Supplementary Fig. 1g) defining areas of interaction. Four disease paradigms were simulated, inspired on real scenarios (a representative reference is given for each paradigm). We next describe what is specific about each paradigm:

**Phenotype Marker Intensity (PMI)**[26]**.** The simulated patient cohorts displayed a 25%, 50%, or 75% relative level of expression of marker Mk6 in cell phenotype Ph6, existing in neighborhood Nb3, for patients of types I, II, and III, respectively (Supplementary Fig. 2). Two experiments with different levels of complexity were performed. In experiment PMI1, the relative abundance of Ph6 cells in Nb3 was set to 15% (moderately present), whereas in PMI2 the relative abundance of Ph6 was set to 0.25% (rarely present).

**Phenotype Frequency (PF)**[16]**.** The simulated patient cohorts displayed a different abundance of cell phenotype Ph6 as a function of patient type. Two experiments were simulated. In PF1 (moderate presence) the relative abundance of Ph6 cells in neighborhood Nb3 was set to 0%, 30%, and 60% for patient types I, II, and III, respectively (Supplementary Fig. 3). In PF2 (rare presence), the relative abundance of Ph6 in Nb3 was set to 0%, 0.12%, and 0.25% for patient types I, II, and III, respectively.

| Patient Types | Metric / Disease Paradigm | PMI1 | PMI2 | PF1 | PF2 | CCI1 | CCI2 | NNI1 |
|---|---|---|---|---|---|---|---|---|
| I, III | Accuracy % ± (CI 95%) | 100±0 | 100±0 | 100.0±0.0 | 93.8±5.3 | 98.8±1.9 | 86.9±7.4 | 90.0±6.6 |
| | Interpretability (%) | 92.4 | 29.0 | 88.2 | 68.4 | 68.5 | 86.5 | 63.8 |
| I, II, III | Accuracy % ± (CI 95%) | 100.0±0.0 | 100±0 | 100.0±0.0 | 70.8±9.1 | 97.5±2.8 | 61.8±10.6 | 75.0±7.7 |
| | Interpretability (%) | 90.8 | 23.6 | 87.5 | 78.3 | 70.3 | 44.1 | 83.5 |
| | Contrast Accuracy (%) | 64.92% | 67.23% | 64.92% | 67.23% | 67.42% | 69.23% | 69.27 |

**Table 1. NaroNet performance: synthetic experiments.** NaroNet's classification accuracy (and 95% confidence interval) and interpretability calculated as the intersection of the most relevant extracted TME and the ground-truth of each synthetic experiment. Legend. PMI: phenotype marker expression; PF: phenotype frequency; CCI: cell-cell interaction, and NNI for neighborhood-neighborhood interaction. Index 1 refers to moderate presence, and index 2 to rare presence.

**Cell-Cell Interactions (CCI)**[17]**.** In these simulated patient cohorts, cell phenotypes Ph4 and Ph5 that coexist in neighborhood Nb2 repel, show no interaction, or attract each other in patient types I, II, and III, respectively (Supplementary Fig. 4). Two experiments were performed (CCI1 and CCI2). In CCI1 (moderate presence) the relative abundance of both Ph4 and Ph5 in Nb2 was set to 5%; in CCI2 (rare presence), the relative abundance of both Ph4 and Ph5 was set to 1%.

**Neighborhood-Neighborhood Interactions (NNI)**[20]**.** The simulated patient cohort displayed different interactions between cellular neighborhoods, related to patient type. In this experiment (NNI1), Nb2 and Nb3 repel, show no interaction, or attract in patient types I, II, and III, respectively. The relative abundance of both Nb2 and Nb3 was set to 15% (Supplementary Fig. 5).

The results, in terms of NaroNet's unsupervised predictive accuracy and interpretability are shown in **Table 1**. Overall, the model predicts remarkably well all disease paradigms, even those involving rare cell populations.

**Illustrative example.** We now illustrate the methodology, accuracy, and interpretability of NaroNet making use of the first experiment of the CCI paradigm (CCI1). The specifics of CCI1 are graphically described in **Fig. 2a,b** and Supplementary Fig. 4:

The Patch contrastive learning (PCL) module (**Fig. 1c**, **Online methods - Patch Contrastive Learning**) was unsupervisedly trained to produce 256-dimensional vector embeddings from 128 randomly located 10x10 pixel image patches. Then, each cohort image was divided into 6,400 10x10 pixel patches and fed to the PCL CNN to produce a vector embedding for each patch. The contrast accuracy provided by the PCL module was 67.42%, (**Table 1**). This value, which evaluates the ability of the resulting embedding to distinguish elements contained within each patch, is similar to the state-of-the-art in semi-supervised learning setups[27]. All patches extracted from each image were assembled into a neighborhood graph (**Fig. 1d**, **Online methods – Patch-Graph generation**), and fed to NaroNet (**Fig. 1e**, **Online methods –NaroNet**). An architecture search module (**Online methods - Architecture Search**) was used to find the model configuration that maximized both interpretability and accuracy. Supplementary Table 1 contains the parameters evaluated and their corresponding search space. Supplementary Table 2 shows the results produced by the architecture search module. To quantitatively evaluate the effect of the parameter choices, we compared the performance obtained with and without each parameter using a Mann-Whitney-Wilcoxon two-sided test (Supplementary Fig. 6 and Fig. 7). The parameters found to be most relevant in this experiment (i.e., with the lower p-values) were the use of patch entropy loss (Supplementary Fig. 6a), ResNet module for the graph convolution unit (Supplementary Fig. 6g), and a softmax patch activation function (Supplementary Fig. 7a). Interestingly, the optimal dimensionality value found for the hidden layer was 64 (Supplementary Fig. 7h), which is low compared to what is common in DL workflows. This is a benefit of the use of the enriched graph of low-dimensional patch embeddings, which reduces the requirement of NaroNet's network depth. As expected, the model's performance varies significantly with the number of TMEs used, being the optimal values: 10 phenotypes (Supplementary Fig. 7c), 9 neighborhoods (Supplementary Fig. 7d), and 8 areas of neighborhood interaction (Supplementary Fig. 7e).

Using the optimal architecture, a 10-fold cross-validation was used to measure the classification performance. The receiver operating characteristic (ROC) curves, confusion matrix, and training and test accuracy curves obtained for this paradigm are shown in Supplementary Fig. 8. As shown in **Table 1**, the overall accuracy achieved was 97.50% with a 95% confidence interval (CI) of [94.7,100], i.e., only 3 patients out of 120 were misclassified.

Then we quantified the interpretability of the model, i.e., the ability to automatically identify which TMEs are most related to the patient-outcome. To this end, for the test fold that provided the highest level of precision, we first generated a global hierarchical heatmap containing the presence of each marker in the 9 cellular neighborhoods learned by our model (Supplementary Fig. 9b). Representative patches assigned to each neighborhood are shown in Supplementary Fig. 9d-l. Next the information of the entire cohort of patients, extracted from the neural activations of NaroNet, was displayed in a hierarchical heatmap that shows, for each patient, its relative neighborhood abundances (Supplementary Fig. 10). As intended, and consistent with the ground truth disease paradigm, the patients were correctly clustered by type using neighborhood abundances. Finally, we evaluated which specific TMEs drove the prediction. Our differential TME composition analysis (**Online methods – BioInsights**) revealed that four learned neighborhoods were most predictive (those with the lowest p-value according to a Mann-Whitman test) (Supplementary Fig. 11): in order of statistical significance, N7, N3, N4, and N6. Indeed N7 (**Fig. 2d,e**) and N4 (Supplementary Figure 11) contain spatially related Ph4 and Ph5 cells (Supplementary Fig. 9g), while N3 (**Fig. 2f,g**) and N6 (Supplementary Figure 11) contain Ph4 and Ph5 cells that are seldom associated (Supplementary Fig. 9i). From a global cohort perspective and consistent with this observation, the boxplots that show the abundance of these four relevant neighborhoods (Supplementary Fig. 11c-f) prove that N7 and N4 are the most abundant neighborhoods in type III patients (attraction), N3 and N6 are the most abundant neighborhoods in type I patients (repulsion), and N3 and N4 are the most abundant neighborhoods in type II patients (no interaction), showing an equilibrium between attraction and repulsion. In summary, we show that the TMEs learned by NaroNet capture the underlying disease paradigm and lead to a successful patient classification.

## Synthetic tissue simulation: Cell cell interaction disease paradigm

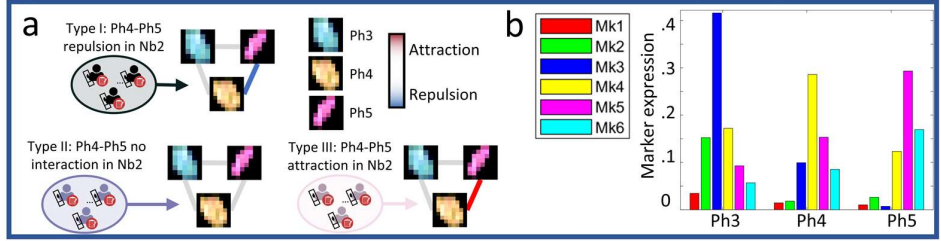

## NaroNet learned TMEs when classifying patient types

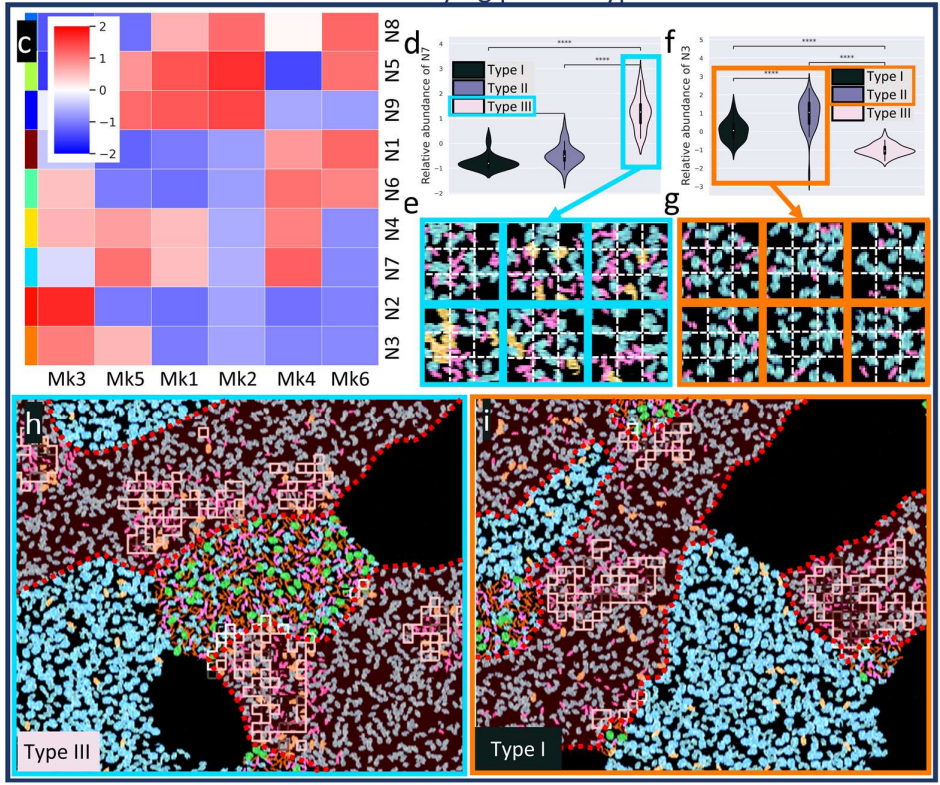

**Fig. 2: Graphical description of synthetic experiment CCI1. a.** Ground truth: Schematic description of the interactions between cell phenotypes Ph3, Ph4, and Ph5 (located in neighborhood Nb2) that define each patient type (I-III). **b**. Marker expression levels for the three relevant cell phenotypes. **c**. Z-scored mean expression of all markers in the neighborhoods learned by NaroNet. **d.** Relative abundance of learned neighborhood N7 in the three patient groups. **e.** Representative patches assigned to N7. **f.** Relative abundance of learned neighborhood N3 in the three patient groups. **e.** Representative patches assigned to N3. **h.** Example of patient correctly classified as type III (i.e. displaying Ph4-Ph5 attraction), with squares showing patches assigned to learned neighborhood N7, located in ground truth neighborhood Nb2 (marked in red). **i.** Example of patient correctly classified as Type I (Ph4-Ph5 avoidance), with squares showing patches assigned to learned neighborhood N3, located in ground truth neighborhood Nb2.   (*** $p < 0.001$; **** $p < 0.0001$)

Finally, we measured (see **Online methods – BioInsights**) the average intersection between the most relevant TMEs found in the images, and the ground-truth neighborhood mask (Nb2) that contains the relevant, patient-type dependent TMEs in this experiment (i.e. Ph4 and Ph5) (Supplementary Fig. 12). The average interpretability for this cohort was 70.3% (**Table 1**), confirming that our model correctly identified the relevant TMEs.

To interpret why an individual image/patient was classified as a certain patient type, we calculated the Predictive Influence Ratio (PIR) value for each TME (see **Online methods – BioInsights**). This strategy applied to CCI1 shows (Supplementary Fig. 12a) that for most type I patients, the abundance of neighborhood N6 was the most determinant classification factor. Conversely, N3 was highly predictive for type II patients, and N4 and N7 were highly relevant to successfully classify type II and III patients. We illustrate this with examples of individual predictions: a patient classified as type III with prediction confidence of 97.75% and a PIR value of 2.28 for N7 (**Fig. 2h**), and a patient classified as type I with prediction confidence of 94.36% and a PIR value of 1.61 for N6 (**Fig. 2i**).

**Experiments with real samples.**

**Endometrial high-grade carcinomas**

Whole slide FFPE tissue sections from 12 patients with high-grade endometrial carcinomas (**Online methods – Endometrial Carcinomas. Histological and Molecular Classification**) were stained using an Opal seven color kit (**Online methods – Endometrial Carcinomas. Tissue Processing**). Images of 336 distinct tumor areas were acquired at a 20x magnification using a Vectra® Polaris$^{TM}$ Automated Quantitative Pathology Imaging System (Perkin Elmer Inc., Waltham, MA, USA). We used NaroNet to learn TMEs associated to four patient-level labels: the somatic POLE mutation, copy number variation, DNA mismatch repair (MMR) deficiency, and two tumor histology types (endometrial carcinoma and serous-like carcinoma).

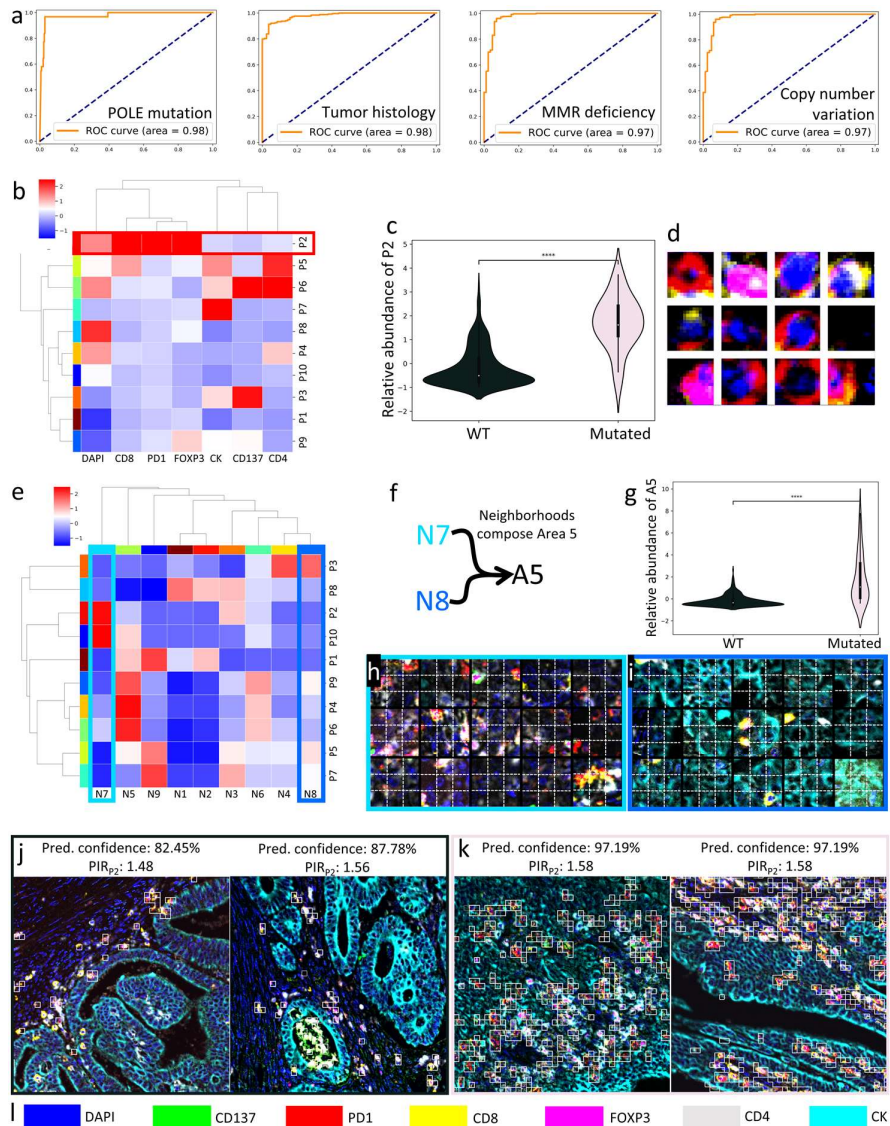

**Fig. 3: Association of high-grade endometrial carcinomas with patient-level labels. a.** ROC curves showing the classification performance of NaroNet for the four tissue characteristics learned. **b.** Heatmap showing the Z-scored mean marker expression, for the phenotypes learned by NaroNet. **c.** Violin-plot showing relative abundance of learned phenotype P2 as a function of POLE mutation status. **d.** Top-12 15x15 patches assigned to local phenotype P2. **e.** Heatmap showing interactions between the local phenotypes learned by NaroNet, and neighborhood composition of learned Area A5. **g.** Violin-plot showing the abundance of A5 as a function of POLE mutation status. **h,i.** Top-15 patches assigned to neighborhoods N7 and N8. **j,k.** Image crops from 2 image fields that were assigned top-2 PIR values for P2 in WT and POLE mutated subgroup, respectively. White squares represent patches assigned to P2. **l.** Color legend used to display image colors.

The PCL module was trained to generate 256-dimensional embeddings of 15x15 pixel image patches obtaining a contrast accuracy of 81.11%. Once trained, it produced 3,906,000 image patch embeddings, for all the images of the cohort. All patches that belong to each image were assembled into a graph of patches and fed to NaroNet. The optimal model parameters were calculated using the architecture search module (Supplementary Table 3). Using this optimal set of parameters, a 10-fold cross-validation was used to validate the ability of NaroNet to learn the four patient-level labels from these images, achieving an overall accuracy of 93.75% with a 95% CI [91.16,96.33] (**Fig. 3a** and Supplementary Fig. 13).

As a proof of principle of the type of analyses that can be performed, we analized the interpretability of the model while predicting POLE mutation status with an accuracy of 96.37%. NaroNet unsupervisedly learned 26 TMEs (Supplementary Fig. 14 and Supplementary Fig. 15). Our differential TME composition analysis (**Online Methods - BioInsights**) revealed that the TMEs that were most predictive of the POLE mutation were local phenotype P2 (p-value: $1.42 \times 10^{-2}$) and area A5 (p-value: $6.50 \times 10^{-3}$). P2 contains cells expressing CD8, FOXP3, and PD1 (**Fig. 3b**). Indeed, these cellular markers are significantly more associated to tumors harboring POLE mutations than to POLE wild type (WT) tumors (**Fig. 3c,d**). This is consistent with the literature, as CD8, FOXP3, and PD1 are inflammation markers, and POLE-mutated endometrial carcinomas, usually with a better prognosis than POLE WT, are described to have large lymphocytic infiltrates[28]. Furthermore Area A5 contains interactions between N7 and N8 neighborhoods (**Fig. 3f,g**) which in turn contain local phenotype interactions between P2-P10 and P3-P5 respectively (**Fig. 3e,h,i**). P3 and P5 express CD137, CD4, CD8, while P2 and P10 are composed of CD8, FOXP3, and PD1 along with some unstained cells. In summary, area A5 contains cellular neighborhoods related to inflammation, and points at the existence of interactions between specific immune phenotypes in POLE vs non POLE mutated cancers that could be further explored, as could be done with other, less statistically significant TMEs selected by NaroNet.

To illustrate the interpretability of our results we provide four examples in which phenotype P2 was the most relevant TME selected by NaroNet. Two of the examples were correctly classified as POLE WT with a prediction confidence of 82.45% and 87.78%, and PIR values of 1.48 and 1.56, respectively. Both show a *cold* tumor landscape –low P2 abundance– that is associated with POLE WT patients (**Fig. 3j**). This is consistent with the global cohort-level findings (**Fig. 3c**). The other two examples were correctly classified as POLE mutated patients with a prediction confidence of 99.54% and 97.19%, and PIR values of 1.58 and 1.40, respectively. Both show a *hot* tumor landscape with high P2 abundance (**Fig. 3k**), being also consistent with global cohort-level findings.

To further validate NaroNet, we quantified the most relevant phenotype found by NaroNet (P2, i.e. high expression of CD8, PD1, and FOXP3) using QuPath[2], a widely used open source software for computational pathology (**Online Methods – Endometrial Carcinoma. Image Analysis using QuPath**). For each image of the cohort, we first quantified the level of expression of CD8, PD1, and FOXP3 from the cell segmentation masks obtained using the DAPI -counterstained- channel. The we calculated the number of $CD8^+PD1^+FOXP3^+$ cells and correlated this number with the number of patches that NaroNet assigned to P2, obtaining positive correlation ($R^2 = 0.72$) (Supplementary fig. 16b). Moreover, the respective violin-plots (Supplementary fig. 16c-d) showed both QuPath and NaroNet are able to robustly distinguish patient-types based on $CD8^+PD1^+FOXP3^+$ phenotype abundance, the outstanding difference being that NaroNet infers it unsupervisedly.

Finally, to test NaroNet's predictive power classifying subjects, not individual images based on the POLE mutation, we performed a leave-one-out experiment -iteratively, 11 patients (represented by all their images) were used to train the model and one patient was used for testing-. Patient-wise predictions were calculated as the mean prediction value of all images that correspond to the test patient, achieving an overall accuracy of 91.67% with a 95% CI [76.04-100.00%], and an AUC of 0.67 with a 95% CI [0.32-1].

**Breast cancer cohort**

NaroNet was trained to predict patient survival risk in a publicly available breast cancer cohort of 283 patients represented by 372 35-plex mass cytometry images[24]. Patients were clustered by k-means into three risk groups (RI, RII, RIII) based on their long-term survival (**Fig. 4a**): RI contained 29 patients that survived more than 120 months, RII contained 107 patients that survived between 54 and 119 months, and RIII contained 48 patients that survived less than 53 months. The PCL module produced 18x18 pixel patch embeddings for all the images of the cohort, with a contrast accuracy of 82.50%. As more than one image was acquired per patient, all the images and patches of each patient were combined in one single graph and fed to NaroNet. The optimal model architecture (Supplementary Tab. 4) was used in a 10-fold cross-validation to classify patients based on their risk survival group. NaroNet predicted RI vs. RIII patients with an accuracy of 71.19% with 95% CI [64.52-77.86] and an AUC of 0.78 with 95% CI [69.93-84.70] (Supplementary Fig. 17).

As an example of how to make use of its interpretability, NaroNet learned 48 TMEs predictive of the patient risk group (**Fig. 4c** and Supplementary Fig. 18). Our differential TME composition analysis revealed that a combination of a phenotype (P6) and a neighborhood (N15) was significantly predictive ($p<0.05$) when distinguishing between RI and RIII patients. N15, a neighborhood that contains high expression of Sox9 and CD20, is more abundant in high risk (RIII) patients compared to low risk (RI) patients (**Fig. 4d**). This is consistent with several studies that describe the oncogenic activity of Sox9[29], even if the role of this gene has yet to be understood when combined with CD20. Moreover, the abundance of Phenotype P6, consisting of ER and HER2 cells, was also associated with a better prognosis (**Fig. 4e**). This is also confirmed by studies (e.g., Parise et. al.[30]) that show that indeed, the combination of ER and HER2 is clinically linked to a better prognosis. To illustrate that even non-statistically significant predictive TMEs could be used to obtain insights in "discovery" mode, we evaluated the presence of other TMEs in all patient types. For instance, Sox9 containing neighborhood N4, was associated with lower survival, and seen for neighborhood N15 (see supplementary Fig. 18b).

We next evaluate NaroNet's ability to capture the heterogeneity of the 35-plex breast cancer cohort by showing the interpretable prediction of two patients. The first patient, with a survival of 174 months, was correctly classified as low risk (Rip) with a prediction confidence of 71.62%. Such prediction was mainly driven by the presence of P6 (i.e., ER, HER2), having a high abundance (5.44) and high PIR value (1.51). This is consistent with our global findings that indicate that low risk patients are associated with the presence of P6. The second patient, with one month survival, was correctly classified as high risk (RIII) with a prediction confidence of 65.76%, being the prediction mainly driven by the presence of N15 (i.e., Sox9 and CD20), having a high relative abundance (1.33) and a PIR value (1.29). Therefore, once more, for a specific patient, NaroNet correctly quantified a high presence of N15, and associated it to low survival (**Fig. 4d**).

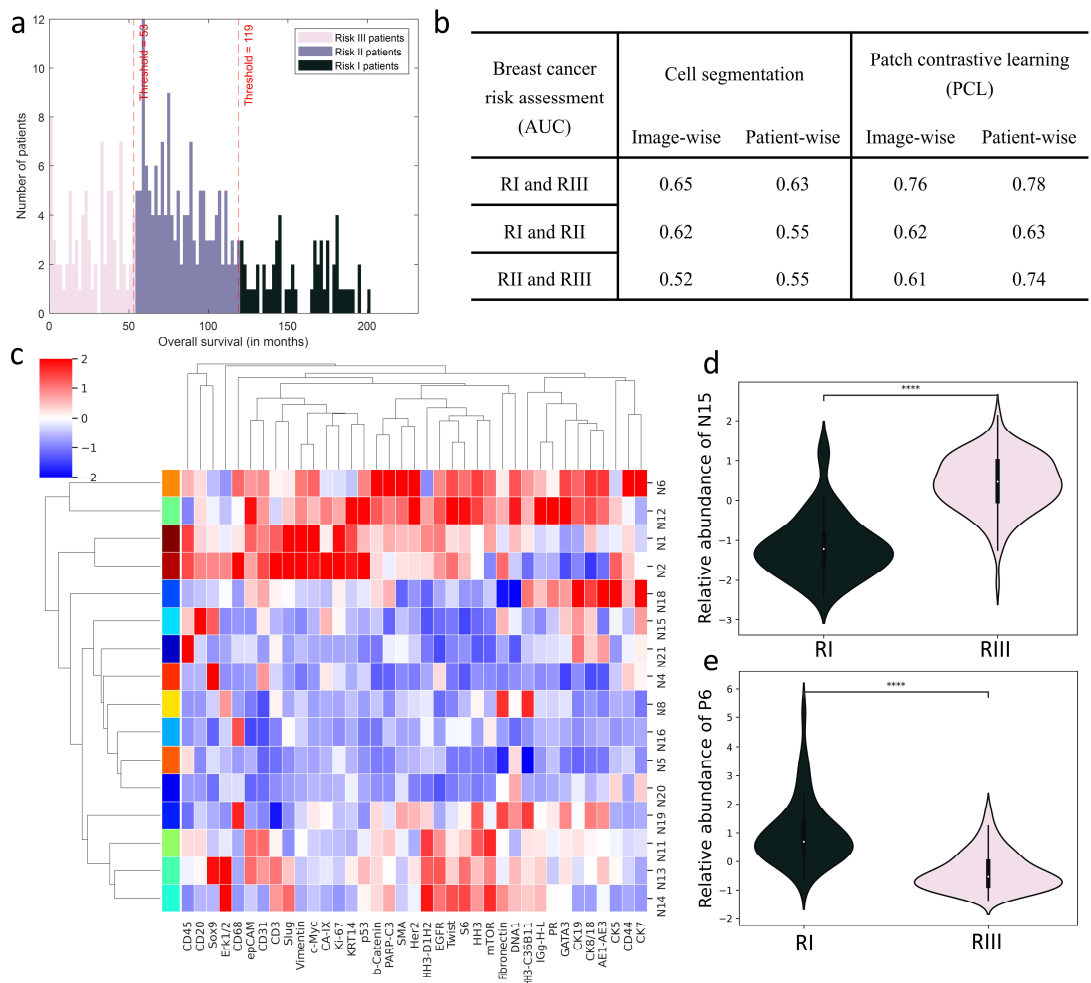

**Fig. 4**: **Association of spatially-resolved 35-plex of breast cancer tissues with the long term survival. a.** Histogram plot of patient overall survival in months, colored by risk classes. **b.** AUC prediction performance of NaroNet for four training strategies: image-wise or patient-wise, and using the cell segmentation provided with the image dataset or using the proposed PCL method. **c.** Heatmap showing the Z-scored mean marker expression, for all neighborhoods learned by NaroNet. **d.** Violin-plot showing the relative abundance of learned neighborhood N15 as a function of risk group (RI and RIII). **d.** Violin-plot showing the relative abundance of learned neighborhood P5 as a function of risk group (RI and RIII).

We finally analyzed NaroNet's ability to predict patient risk subtypes as a function of the input used. On the one hand, as cell segmentation masks are provided along with the public image dataset, cell features were extracted as done by Jackson et al.[24]. Briefly, a graph of interconnected cells (37-element vectors) was fed to NaroNet, where each cell vector consists of the average expression of the 35 markers plus the cell size and eccentricity. This approach is compared to our proposed strategy based on the use of graphs of patch embeddings. On the other hand, as more than one image was acquired for some patients, we compared the strategy of feeding NaroNet using one graph per image or feeding it with a graph that combines all the images of the patient. For each experiment, architectural parameters were optimized (see supplementary Tab. 4) and were then used in 10-fold cross-validation to classify patients/images concerning their risk survival. **Fig. 4b** shows the area under the curve (AUC) for all experiments. As shown, NaroNet achieves the highest prediction performance using graph containing PCL patches instead of cell masks, and does it better when all images of the same patient are combined into a single graph.

## Discussion

Our hypothesis is that TMEs can be *blindly* identified and associated with patient-level tumor information from multiplex imaging data. We have presented NaroNet, an end-to-end machine learning framework that proves this hypothesis true, as it accurately performs patient predictions from local phenotypes, neighborhoods, and neighborhood interaction areas, unsupervisedly identified from multiplex histological data. NaroNet takes advantage of, and improves elements of the two main state-of-the-art computational pathology approaches. From SCA methods, NaroNet inherits the use of graphs to capture phenotype interactions, extending this idea by using GNNs to *learn* a three-layer model of the tumor microenvironment from patient-level labels only. From WSDL, NaroNet uses the concept of using patches instead of cell segmentation masks and applies it to multiplex immunostained sections instead of H&E histological images. Furthermore, instead of being a black-box approach[31], NaroNet is inherently interpretable as it makes predictions based on TME annotations, where multiplex data of large patient cohorts can be easily assessed in heatmaps and linked back to the images. This is especially appropriate for the discovery of biomarker signatures.

Besides those main conceptual contributions, NaroNet's specific methodological contributions are numerous. The most relevant are: i. The use of patch contrastive learning to generate low dimensional embeddings[27] that maximize the prediction performance when the number of patients is low. ii. The use of graphs provides unprecedented flexibility to represent TME interactions and allows combining several images from the same patient into a single graph providing improved predictions. These sparse graphs reduce memory usage by ~4 orders of magnitude, allowing training NaroNet with large patient graphs in GPUs. iii. The use of adjustable graph data augmentation artificially increases the pool of patients. iv. The use of a global relation reasoning unit (GloRe) that integrates information from disjoint and distant regions as well as from different tissues/patients, thus learning some technical variabilities caused by the presence of auto-fluorescence, spectral leakage between biomarkers, low expression of some antigens, etc. v. The use of trainable matrices to assign patches to TMEs, a max-sum pooling operation to represent patients as vectors of abundances, and a patch entropy loss to regularize the confidence by which patches are assigned to specific TMEs. These choices render NaroNet inherently interpretable. vi. The postprocessing use of differential TME analysis to infer relevant TMEs, and a novel PIR metric to reveal the influence of specific TMEs on individual predictions.

We have validated NaroNet using both synthetic and real data. Using a novel multiplex tissue image simulator we created realistic patient cohorts with tunable presence of specific TMEs, providing an ideal objective benchmark to test the performance of the system. Indeed, our extensive validation using synthetic data successfully confirms that NaroNet can learn relevant TMEs –local phenotypes, cell-interaction neighborhoods, and neighborhood-interaction areas–, even when their presence in the tissue is rare. Using a high-grade endometrial carcinoma patient cohort, NaroNet found, among other TMEs that could be explored, a local phenotype expressing CD8, PD1, and FOXP3 whose high abundance was associated with the POLE mutation, while achieving a prediction accuracy of 91.67%. This finding is in accordance with what has already been described in the literature. Moreover, we confirmed using a semi-automated computational pathology software (QuPath), that the abundance of this specific phenotype correlates positively with the one found by NaroNet. This nicely shows that NaroNet can be a useful tool in research environments, as it can help to *blindly* identify novel TMEs that are related to the biology of the tumor. Using a public breast cancer dataset, NaroNet found TMEs that were associated with the patient's survival achieving an AUC of 0.78. Strikingly, NaroNet did not require human supervision to learn, from a pool of millions of cells stained by 35 markers, and among other TMEs, a decisive phenotype consisting of cells expressing ER and HER2, and related it to the prognosis of the patient.

In addition, we show that NaroNet performs better when fed with an enriched graph created from image patches than when using cell features obtained from cell segmentation masks. This shows that NaroNet is able to learn relevant tumor microenvironmental information without the highly demanding task of segmenting all cells in the tissue. We also prove that using graphs to represent patients is more advantageous than using images, as patient's stained tissue sections can be combined together into a single graph providing NaroNet with more information to make better predictions.

In summary, we have shown that NaroNet unsupervisedly identifies and annotates the relevant TMEs that truly drive patient-outcomes. Moreover, since NaroNet is able learn *in situ* highly predictive TMEs that confirm the existing literature, it is possible to affirm that the analysis of those predictive TMEs discovered by NaroNet could provide novel insights into the mechanisms of disease progression. This could be used in clinical settings to confirm the predictive value of already known elements of the tumor microenvironment and, more importantly, it makes NaroNet a valuable research tool for the discovery of novel biomarkers. Furthermore, the fact that NaroNet's clinical predictions are directly based on the annotations of TMEs results in an important breakthrough in computational pathology, as it contributes to *whitening* DL black-boxes. Indeed, our model allows clinicians to understand which TMEs drive the prediction of each patient safely and reliably since DL neuron activations are related to specific biological structures that can be mapped back into the original images. Therefore, NaroNet could be an optimal solution for the rapid clinical translation of biomarker discovery signatures, where DL models trained to quantify relevant TMEs are then applied to new incoming patients by providing clinicians with interpretable predictions.

# Online methods

## Patch Contrastive Learning

PCL is inspired in the state of the art in image classification in situations of limited availability of annotations, or *semi-supervised* learning[27,32]. These methods learn enriched image representations from large numbers of unlabeled images using an unsupervised deep neural network. Then, a simple classifier can be trained to obtain image predictions from a small subset of these enriched image representations, in a supervised way. In our case, we use PCL to create, in a non-supervised way, low-dimensional enriched representations of tumor microenvironment regions (or patches) from multiplex immunostained images.

Our PCL strategy starts by dividing the tissue images in patches of a size that could include between none and two cells and contain, for each pixel, the spectral information provided by the acquisition system. These patches are our basic unit of representation of the tissue microenvironment. As illustrated in (Supplementary Fig. 19), our PCL module learns an embedding of these patches by maximizing agreement between different augmented views of the same image patch using a contrastive loss function. To this end, every image channel of the cohort is first Z-score normalized to account for biomarker intensity differences across the entire patient cohort. Then, the following steps are applied iteratively to train the PCL module:

- **Image crop selection.** A set $\{x\}_{1..B_{CR}}$ is created made of $B_{CR}$ image crops of size $(S_{CR} * \alpha_{CR}) \times (S_{CR} * \alpha_{CR}) \times B$ obtained at random positions from a set of R random images chosen from the pool of $I$ existing images (Supplementary Fig. 19a-b). The choice of $S_{CR}$ is critical as it determines the extent to which biological structures can be captured, and their context. It also determines the size of the graph that is used to predict the outcome of the patient (see next section). This value was chosen considering that: (i) a patch should be only large enough to contain zero, one or two cells to ensure interpretability at the level of single cell or small cell environments, and (ii) the number of patches extracted from the images should fit in a single GPU to produce patient predictions efficiently.

- **Data augmentation.** Data augmentation is used to generate two augmented views, or patches, $\tilde{x}_{j1}$ and $\tilde{x}_{j2}$, from each image crop $x_j$ that belongs to the set $\{x\}_{1..B_{CR}}$. This way we obtain an image patch set $\{\tilde{x}\}_{1..B_{CR}*2}$. The data augmentation module (Supplementary Fig. 19c) applies the following sequence of simple transformations to each image crop $x_j$: i. two *random crops* to obtain two image patches $\tilde{x}_{j1}$ and $\tilde{x}_{j2}$ of size $S_{CR}$; ii. one *random rotation* both to $\tilde{x}_{j1}$ and $\tilde{x}_{j2}$; and iii. a *random cutout* to both $\tilde{x}_{j1}$ and $\tilde{x}_{j2}$ consisting of masking out random sections of the patch of size $0.15 \times S_{CR}$. Note that we do not include augmentations that cause color distortions. The reason is that the spectral information is based on immunofluorescence marker expression, and adjustments of these values could change their biological meaning.

- **Convolutional Neural Network.** The set of augmented patches $\{\tilde{x}\}_{1..B_{CR}*2}$ is fed to a ResNet-101 (Supplementary Fig. 19d), to obtain a low dimensional vector representation or embedding of the patches $\{h\}_{1..B_{CR}*2}$, being each patch, $h_{jk} = ResNet(\tilde{x}_{jk}), k = 1,2$ where $h_{jk} \in \mathbb{R}^g$. We chose the dimensionality of the new set $g$=256. Then, a Multilayer Perceptron (MLP) maps each representation $h_{jk}$ to a 128-dimensional vector $z_{jk}$. To this end, the MLP has two layers, containing one hidden layer $z_{jk} = MLP(h_j) = W^{(output)}\sigma(W^{(hidde\ )}h_{jk})$ where $W$s are trainable parameter matrices, and $\sigma$ is a ReLU activation layer.

- **Contrastive loss.** Finally, a contrastive loss function is applied to $\{z\}_{1..B_{CR}*2}$ to create similar embeddings for patches contained in the same crop (i.e, $z_{j1}$ and $z_{j2}$) -possibly corresponding to the same biological structure-, while forcing dissimilar embeddings for patches contained in different image crops (i.e. $z_{ik}$ and $z_{ql}$, being $i \neq q$) -possibly corresponding to dissimilar biological structures- (Supplementary Fig. 19h-i). Let $sim(u,v) = u^T v / \|u\| \|v\|$ denote the cosine similarity between two vectors $u$ and $v$. To this end, the loss function for any given pair of patches that belong to the same image crop is defined as:

$$\ell_{j1,j2} = -\log \frac{exp(sim(z_{j1}, z_{j2})/\tau)}{\sum_{q=1, q \neq j, l=1,2}^{B_L - 1} exp(sim(z_{jk}, z_{ql})/\tau)} \qquad (1)$$

where $\tau$ is a temperature parameter set to 0.5.

Once the PCL module is trained, it is used to infer image representations from all the patches of the images of the cohort. Given a high dimensional multiplex image $i \in \mathbb{R}^{i_x \times i_y \times B}$, this is divided into a grid of evenly distributed patches of size $S_{CR} \times S_{CR} \times B$. Each image patch is then introduced into the PCL module to obtain an image patch representation $h_j$ (Supplementary Fig. 19d). Therefore, when every patch of the image $m$ is extracted, the resulting image has a reduced dimensionality, i.e. $i \in \mathbb{R}^{i_x \times i_y \times B} \rightarrow \mathbb{R}^{L \times g}$, where $L = \frac{i_x i_y}{S_L^2}$ is the number of patches of the image, and $g$ is the number of features contained in the new patch embedding, in our case 256. Therefore, for each image, the PCL module outputs a list of patches with their respective representations $\{h\}_{1..L}$, and their corresponding position on the image. All in all, the PCL module both generates useful visual representations and reduces image data size by approximately one order of magnitude.

**Patch-Graph generation**

We model the tissue using three levels of information corresponding to three types of TMEs (i.e., tissue phenotypes, phenotype neighborhoods, and neighborhood interactions). To model each tissue at the lowest level of tissue phenotypes, we use the set $\{h\}_{1..L}$ of patch representations that are provided by the PCL module. To capture phenotype neighborhoods, i.e., tissue phenotypes that are spatially associated, we create a graph that contains all the patches of the tissue, by connecting each image-patch to its 4 adjacent neighbors. This graph is $\mathcal{G} = (Z, A)$, where $Z \in \mathbb{R}^{L \times g}$ is a matrix that contains all the embeddings of the image patches $\{h\}_{1..L}$, and $A \in \{0,1\}^{L \times L}$ is an adjacency matrix that contains the connectivity between patches. To reduce the expensive memory burden of storing complete adjacency matrices we 'sparsify' $A$ into $A' \in \mathbb{Z}^{E \times 2}$, generating $A'$ as a list of edges (i.e., connections) between patches, extracted from the non-zero values of the original $A$, where $E$ is the number of edges present in the graph (i.e., the positive values from $A$). Thus graph $\mathcal{G} = (Z, A)$, is converted into graph $\mathcal{G}' = (Z, A')$. Since we connect each patch to its 4 adjacent neighbors, i.e., $E = L \times 4$, the memory required to store $A'$ increases linearly compared to $A$, for which it increases exponentially. For example, in a case where $L = 10,000$ patches, the memory required to store a patient's tissue information is reduced by ~4 orders of magnitude.

**NaroNet**

Having a cohort of patients $\mathcal{D} = \{(G_1, y_1), (G_2, y_2), \ldots, (G_M, y_M)\}$, where $M$ is the number of patients, each patient represented by a graph $G_m \in \mathcal{G}$, and a patient-level label $y_m \in \mathcal{Y}$, the goal of NaroNet is to learn a mapping $\mathcal{G} \xrightarrow{f} \mathcal{Y}$ that connects patient information to patient predictions. To this end, the architecture of NaroNet is divided into two consecutive sections $\mathcal{G} \xrightarrow{f_1} (\mathcal{P}, \mathcal{N}, \mathcal{A}) \xrightarrow{f_2} \mathcal{Y}$, trained end-to-end using the patient labels, where $\mathcal{P}$ is the abundance of tissue phenotypes $\mathcal{P} \in \mathbb{R}^P$, $\mathcal{N}$ is the abundance of neighborhoods $\mathcal{N} \in \mathbb{R}^N$, and $\mathcal{A}$ is the abundance of areas of interaction between neighborhoods $\mathcal{A} \in \mathbb{R}^A$. For the sake of consistency, we refer globally to $\mathcal{P}, \mathcal{N}, \mathcal{A}$ as tumor microenvironment elements (TMEs). P, N, and A are respectively the number of phenotypes, neighborhoods or areas or neighborhood interactions.

The first section of NaroNet, $f_1$, is an ensemble of three parallel networks whose objective is to assign nodes to distinct $\mathcal{P}, \mathcal{N}, \mathcal{A}$ values, and the second one, $f_2$, is dedicated to obtaining patient's predictions from the tumor microenvironment. To learn the tumor microenvironment, three neural networks $\boldsymbol{f_1} = (f_{1P}, f_{1N}, f_{1A})$ are trained in parallel from individual patient data and later pooled to obtain the abundance of each TME, as described in the following paragraphs:

- **Assignment of patches to phenotypes.** Each image patch, $h_l \in Z_m$, is assigned to a phenotype vector using $f_{1P}$, i.e.:

$$S_P = f_{1P}(Z_m) \in \mathbb{R}^{L \times P}, \qquad (2)$$

where $f_{1P}$ is an 8-layer MLP with skip connections, and $P$ elements/nodes in the last layer. Therefore $f_{1P}$, takes the node embeddings belonging to image $Z_m$ and provides a node assignment matrix $S_P$.

- **Assignment of patches to neighborhoods.** Likewise, each image patch, $h_l \in Z_m$ is assigned to a neighborhood $f_{1N}$, i.e.:

$$S_N = f_{1N}(Z, A') \in \mathbb{R}^{L \times N}, \qquad (3)$$

where $f_{1N}$ is a Graph Neural Network (GNN) followed by a 1-layer MLP which has a dimensionality $N$. Therefore, $f_{1N}$ uses the node embeddings of $Z_m$ and the adjacency matrix $A'$ to produce a node assignment matrix $S_N$. GNNs are neural networks that capture interactions and relationships between connected nodes of a graph. To that end, it iteratively performs a trainable weighted sum of the feature vectors from each graph node (in our case patch $h_l \in Z_m$) and its connected neighboring nodes, generating a new feature vector at the next hidden layer of the network[33,34,35]. Here we use the following GNN formulation:

$$Z^{(k)} = ReLU(A'Z^{(k-1)}W^{(k-1)}) \in \mathbb{R}^{L \times H} \qquad (4)$$

where $Z^{(k)}$ are the node embeddings in the $k^{th}$ hop, and $W^{(k)} \in \mathbb{R}^{H \times H}$ is a trainable weight matrix, being $H$ the dimensionality of the output feature vector. A full GNN module will run $K$ iterations of eq. (4) to generate the final output embeddings $Z^{(K)} \in \mathbb{R}^{L \times H}$, where $K$, also referred to as the number of hops of the GNN, represents the extent to which the output patch embeddings can capture information of their neighbors.

- **Assignment of patches to neighborhood interactions.** Each neighborhood patch resulting from the previous GNN, $h_l \in Z_m^{(K)}$ is assigned a neighborhood interaction patch using a second GNN ($f_{1A}$). To this end the following trainable assignment matrix is used:

$$S_I = f_{1A}(S_N^T Z^{(K)}, S_N^T A' S_N) \in \mathbb{R}^{N \times A}, \quad (5)$$

This GNN learns the interactions between N neighborhoods, therefore learning higher-order interactions of the original graph. For this purpose, the inputs for $f_{1A}$ are the embeddings of the $N$ neighborhoods ($S_N^T Z^{(K)} \in \mathbb{R}^{N \times H}$) and the interactions between neighborhoods ($S_N^T A' S_2 \in \mathbb{R}^{N \times N}$). This way $f_{1A}$ accumulates feature vectors of neighborhoods that are close to each other. As in the previous section, the number of hops of the GNN ($K$), determines the extent to which the patch embeddings can capture information of their neighbors.

Summarizing, after applying $f_{1P}, f_{1N}, f_{1A}$, each row of $S_P$ contains the probability of each patch of the image to contain each of the $P$ phenotypes, each row of $S_N$ contains the probability of a patch of the image to contain each of the N neighborhoods, and each row of $S_A$ contains the probability of a neighborhood of the image to contain each of the possible $A$ areas (Supplementary Fig. 20). Then, the final step of $f_1$ is a max-sum pooling operation that captures the abundance of each TME:

$$\mathcal{P} = \sum_{1..L} \max_{1..P}(softmax(S_P)) \in \mathbb{R}^P. \quad (6)$$

$$\mathcal{N} = \sum_{1..L} \max_{1..N}(softmax(S_N)) \in \mathbb{R}^N, \quad (7)$$

$$\mathcal{I} = \sum_{1..L} \max_{1..A}(softmax(S_A)) \in \mathbb{R}^A. \quad (8)$$

where $S_P, S_N, S_A$ of eq. (6, 7, 8) are the assignment matrices whose values correspond to neuron activations (Supplementary Fig. 20a) and the *softmax* activation function transforms neuron activations into probabilities in a row-wise fashion (Supplementary Fig. 20b). The *max* operator function is applied row-wise so that only the maximum values of each row are kept, while the others are set to zero (Supplementary Fig. 20c). The *sum* operator is applied column-wise to obtain the abundance of each TME. The resulting $(\mathcal{P}, \mathcal{N}, \mathcal{A})$ are the TME abundances that represent each patient.

The TME abundance $(\mathcal{P}, \mathcal{N}, \mathcal{A}) \in \mathbb{R}^{P+N+A}$ vector is fed to an MLP of 1 layer ($f_2$) so that $y' = f_2(\mathcal{P}, \mathcal{N}, \mathcal{A}) \in \mathbb{R}^O$, where $y'$ is the prediction between $O$ possible patient-outcomes. A cross entropy loss is used to train the parameters of both $f_1$ and $f_2$.

The strategy used to implement $f_1$ can produce spurious local minima when all patches are assigned to a single microenvironment element. This locally optimal solution traps the gradient-based optimization. To prevent this we propose the use of three regularization loss functions:

- A **patch entropy loss** is used to regularize the probabilities given by eq. (6, 7, 8), where patches are assigned softly to TMEs. Upon initialization, the assignment of patches to TMEs is uncertain and the entropy of a patch is high. During the training process, we aim at knowing the assignment of patches to TMEs, thus obtaining a sparse matrix assignment. To this end, we propose to reduce patch entropy for each TME using a loss function:

$$\ell = \frac{1}{L} * \sum_{l=1..L} -sum(softmax(S) * \log(softmax(S))) \qquad (9)$$

where $S$ is any of the matrices $(S_P, S_N, S_A)$, and the function generates $(\ell_{ep}, \ell_{en}, \ell_{ea})$ losses, respectively. The final loss is restricted to $\mathbb{R} \cap [0,1]$ where the lower the value the most certain it is that the patch belongs to a specific TME. The final, combined loss is regularized by a $\lambda$ parameter.

$$\ell_e = (\lambda_{ep} * \ell_{ep} + \lambda_{en} * \ell_{en} + \lambda_{ei} * \ell_{ei})/3 \qquad (10)$$

where $\lambda_{ep}, \lambda_{en}, \lambda_{ei}$ regularize how much the weights are adjusted to each TME entropy. This is a specific learning rate that is chosen based on the tumor microenvironment complexity.

- An **orthogonal loss** function is used to avoid graph pooling collapse in GNNs, i.e., to assign every node in a graph to a single cluster[36]. This loss term is an orthogonal regularization term[37] that forces clusters to have the same size, thus containing a similar number of nodes. We adapt this loss to regularize assignment matrices $(S_N, S_A)$.

- A **patient entropy loss** is proposed to avoid graph pooling collapse in $(\mathcal{P}, \mathcal{N}, \mathcal{A})$ TMEs. This strategy is less restrictive than the orthogonal loss since the regularization term does not force clusters to have the same size. Instead, we avoid low entropy of $(\mathcal{P}, \mathcal{N}, \mathcal{A})$ TME abundance vectors, where a value of entropy zero means that all patches are assigned to a single cell type:

$$\ell_{pp} = sum(\mathcal{P} * \log(\mathcal{P})) \quad (11)$$

$$\ell_{pn} = sum(\mathcal{N} * \log(\mathcal{N})) \quad (12)$$

$$\ell_{pa} = sum(\mathcal{A} * \log(\mathcal{A})) \quad (13)$$

where $(\mathcal{P}, \mathcal{N}, \mathcal{A})$ are the TME abundances visually represented in Supplementary Fig. 20d and $(\ell_{pp}, \ell_{pn}, \ell_{pa})$ are restricted to $\mathbb{R} \cap [-1,0]$ where the lower the value the most spread out is the abundance of TMEs. As done for the patch entropy loss, the final loss is also regularized using a $\lambda$ parameter, $\ell_p = (\lambda_{pp} * \ell_{pp} + \lambda_{pn} * \ell_{pn} + \lambda_{pa} * \ell_{pa})/3$, where $\lambda_{pp}, \lambda_{pn}, \lambda_{pa}$.

**Architectural variations**

From the previously explained baseline NaroNet architecture, some variations have been proposed and implemented, after a process of architecture optimization described in the next section:

- **Multiple TME assignment.** In classical computational biology, a cell or tissue element is commonly assigned to a single TME[19,38]. Here we allow patches to be assigned to more than one TME, e.g. patches containing elongated Ki67 positive cells can be assigned to two phenotypes, the first capturing cells with an elongated morphology, and the second cells expressing Ki67. To this end, we eliminate the max operator in eqs. (6, 7, 8).

- **Leveraging patch relevance.** Following the strategy behind standard semi-quantitative analysis of immunostained sections, such as the H-score[39], where cells or tissue elements are assigned a semi-quantitative factor to measure the abundance of TMEs, we also allow our neural networks $f_{1P}$, $f_{1N}$, $f_{1A}$ to assign different weights to each patch in a given TME. To this end, we replace the *softmax* operator in eq. (6, 7, 8) with a *sigmoid* operator. This way neural activations in eq. (6, 7, 8) may add up to values higher to 1, thus eliminating the limitation imposed by the *softmax* operator.

- **Global reasoning unit.** Multiplexed image staining quality can be affected by many factors: the presence of auto-fluorescence, spectral leakage between different biomarkers, low expression of some antigens, etc., leading to a great deal of variability within and between samples. When hundreds of samples are available, it is possible to 'learn' some of the technical variabilities present in the tissue. Here, we use a global relation reasoning unit (GloRe)[40] to capture relations between disjoint and distant regions. To this end, we preprocess the input of NaroNet, patch embeddings are fed to this processing unit, that returns the patch embeddings with the same shape, but with its features being leveraged by the rest of the tissue. This way, the processing unit can use distant regions and normalize their values.

- **Alternative GNN architectures.** Several GNN architectures have been proposed such as residual networks (ResNet)[41], jumping knowledge networks (JKNet)[42], and IncepNet[43]. We have compared these three alternative architectures to determine the one that provides better learning performance.

- **Data augmentation for Graph Neural Networks.** Data augmentation expands the training dataset to improve the model's generalization. Given a graph $\mathcal{G} = (Z, A)$, augmentation operations are carried out sequentially. We investigate which approaches provide a better generalization of patient-outcome predictions and include a tunable parameter $\rho$ that weighs the number of edges/nodes that are being augmented. The most commonly used augmentation operation in GNNs is to drop $\rho \times L$ edges randomly in each training epoch[44]. We also propose to add $\rho \times L$ edges randomly to the graph. Additionally, we mask out node embedding information from $\rho \times L$ nodes.

**Architecture search**

To determine the configuration that provides the highest predictive and interpretability performance, we incorporate an architecture search algorithm to the training phase of NaroNet[45]. This is feasible since, by our choice of using PCL, the large multispectral input images are embedded in compact, low dimensional enriched patch-graphs. Due to this, NaroNet does not require very deep, multilayer neural networks to successfully classify patients. This reduces drastically the training time and allows an architecture search to iteratively optimize the parameters of the network based on the training data.

We choose the Tune framework to perform architecture search[46], which requires defining three main modules (Supplementary Fig. 21)[47]:

- A **search space**, which in our case is the set of interchangeable modules (see Architectural variations), the loss functions, and hyperparameters.

- A **search strategy** used to explore the search space, i.e., ASHA[48], which works by the repetition of three simple steps: (i) train a considerable number of architectures with a few numbers of epochs, (ii) evaluate all configurations, (iii) keep the top $\eta$ architectures, (iv) increase the number of epochs to train the architectures, (iv) and repeat the process until only one architecture is left.

- A **performance estimation strategy** which in our case is based on the classification accuracy of the model on unseen data -valid both for synthetic and real data-, and the interpretability of the model, only for the synthetic data for which an absolute ground truth exists. To this end, the network is trained using a random set containing 90% of the data, and the performance is evaluated on the test set that contains the remaining 10% of the data (one fold from a 10-fold cross validation configuration).

**BioInsights**

Besides generating predictions, NaroNet can be used to study the elements of the tumor landscape that relate to specific predictive tasks. This is done through the quantification of the abundances of the TMEs ($\mathcal{P}, \mathcal{N}, \mathcal{A}$), obtained by NaroNet while classifying patients. The goal of the BioInsights module (Supplementary Fig. 22) is the identification of cohort-differentiating features. This can help answering clinically relevant questions that guide patient prediction: i.e. classification performance (Supplementary Fig. 22b), outlier detection (Supplementary Fig. 22c), differential TME analysis (Supplementary Fig. 22d-e), patient heterogeneity quantification (Supplementary Fig. 22f), as described in the following paragraphs:

- **Patient classification** (Supplementary Fig. 22b-c) NaroNet's classification output is given as a probability value between 0 and 1 for each class. This prediction probability indicates to which extent the patients were correctly classified. Apart from the probability maps given by NaroNet, the BioInsights module provides confusion matrices, ROC, and precision-recall curves, that reflect the global performance of the model.

- **Differential TME analysis** (Supplementary Fig. 22d) NaroNet's $f_2$ network maps TME abundances to patient-outcomes, i.e., $\mathcal{Y} = f_2(\mathcal{P}, \mathcal{N}, \mathcal{A})_{1..M}$. Therefore, $(\mathcal{P}, \mathcal{N}, \mathcal{A})_{1..M}$ are the coefficients or covariates of the model, and the patient's predictions are made solely using the relative abundance of specific TMEs. We use regression analysis[49,50] to interrogate which TMEs were more important to perform patient predictions, i.e., to determine if the abundance of a specific TME is related to disease progression. Specifically, to evaluate whether a specific TME is significant to perform patient prediction, a leave-one-out strategy is used, where a TME $t$ is extracted from the set of all patient TME abundances $(\mathcal{P}, \mathcal{N}, \mathcal{A})_{1..M}$ obtaining a new set of TMEs defined as $(\mathcal{P}, \mathcal{N}, \mathcal{A})_{1..M}^t$. The model is evaluated with the whole patient cohort, and new prediction probabilities are obtained. Then, a Kruskal-wallis test is used to compare the prediction performance of the original TMEs with the leave-one-out model. If the null hypothesis is accepted, the extracted TME is considered to have predictive value.

- **Predictive Influence Ratio (PIR)** (Supplementary Fig. 22e) The above-explained differential TME analysis is useful to find global patterns in patient cohorts but fails to address the heterogeneity existing between patients/tissues[24]. Neural networks like $f_2$ can learn different mechanisms for the same patient type, e.g. in some breast cancer patients the abundance of HER2 protein may be indicative of prognosis, while in other patients, the prognosis could be related to HR$^+$ enrichment[51]. To address this patient-level cohort heterogeneity we introduce the predictive influence ratio (PIR), which quantifies the influence that each TME has on the prediction accuracy. For each patient $m \in \{1 - M\}$:

$$PIR_{m,t} = \frac{f_2(\mathcal{P}, \mathcal{N}, \mathcal{A})_m}{f_2(\mathcal{P}, \mathcal{N}, \mathcal{A})_m^t} \tag{14}$$

where $PIR_{m,t}$ is the predictive influence ratio for a patient $m$ and a TME $t$, and $f_2(\mathcal{P}, \mathcal{N}, \mathcal{I})_m^t$ is the leave-one-out model performance. The higher the value of $PIR_{m,t}$ the most important the TME $t$ is for the classification of patient $m$. Since a dataset is composed of $M$ patients and $T = P + N + A$ TMEs, we can analyze the reason why each patient has been assigned a given prediction.

- **Patient subcategories** (Supplementary Fig. 22e) Once patients' PIR are calculated for each TME, i.e., $PIR_{1..M,1..T}$ we can group patients that have similar disease mechanisms, or in other words, patients that were assigned to a certain class thanks to the abundance of the same TME or the same combination of TMEs. This can help in finding differences within patient cohorts. To this end, we use a hierarchical clustering (fcluster in scipy v1.5.4) that fits the best number of clusters hierarchically.

**Interpretability validation.** In those cases when ground-truth masks are available (e.g. synthetic experiments) referring to regions driving the disease paradigm, we evaluate whether the proposed method classified patients correctly based on them or not. First, we select the TME with the higher PIR value for each image/patient, i.e. most relevant TME, and generate two binary images. The first image, $I_{TME}$, is an image with ones in those pixels that correspond to a patch that was assigned to the most relevant TME. The second, $I_{GT}$ is the ground-truth mask, containing ones in those regions that drive the disease paradigm. Next, we calculate the intersection between both images, and divide the result by the number of pixels present in $I_{TME}$:

$$interpretability = \frac{I_{TME} \cap I_{GT}}{sum(I_{TME})} \quad (15)$$

This way, we calculate the ratio of pixels that were relevant in NaroNet's prediction that lied in the regions of the ground-truth mask that actually contain the disease paradigm. Here, an interpretability value of 1 means that NaroNet's most relevant TME is totally contained in the ground truth mask. This process is repeated for each image/patient in the cohort providing a mean value of interpretability.

**Endometrial Carcinomas. Histological and Molecular Classification**

Formalin-fixed, paraffin-embedded (FFPE) samples from a retrospective collection of 12 high-grade endometrial carcinomas were analyzed, including 9 FIGO grade 3 endometrioid carcinomas and 3 serous-like carcinomas. This collection has been previously reported[52] and included samples seen at the Instituto de Investigación Biomédica de Bellvitge (IDIBELL, Lleida, Spain). All tissue was used after approval from the IDIBELL Investigation Committee protocol numbers: CEIC-2083, which approved the patient consent forms. All cases in the cohorts were reviewed by a local pathologist using hematoxylin and eosin–stained preparations and the tumor histology variant was confirmed by morphology analysis. All cases have been molecularly classified in previous studies[52]. The microsatellite instability (MSI) subclass was determined by immunohistochemistry for DNA mismatch repair (MMR) proteins. Genomic copy number variation was determined using an array comparative genomic hybridization (aCGH) on sample pairs as previously described[53]. POLE variants were studied by Sanger sequencing or NGS of the complete exonuclease domain (exons 9–14) or targeted sequencing of exons 9, 13, and 14.

**Endometrial Carcinomas. Tissue Processing**.

The exploration of the immune environment of high-grade endometrial carcinomas was done using a previously developed and validated seven-color multiplex panel targeting CD4+ and CD8+ T cells, the transcription factor FOXP3, the bona fide T cell activation marker CD137 (4-1BB), the programmed cell death-1 (PD-1), cytokeratin, and nuclear detection with DAPI . The multiplex immunofluorescence panel was used to enable the simultaneous examination of several cellular markers on a single FFPE tissue section. Multiplex immunofluorescence development and validation workflow and protocols have been implemented as previously described[54,55,56].

Briefly, 5 μm sections of FFPE tissue were deparaffinized and antigen retrieval was performed using DAKO PT-Link heat induced antigen retrieval with low pH (pH6) or high pH (pH 9) target retrieval solution (DAKO). Each tissue section was subjected to five successive rounds of antibody staining, each round consisting of protein blocking with antibody diluent/block (Akoya Biosciences ARD1001EA), incubation with primary antibody, Opal Polymer anti mouse/rabbit HRP (Akoya Biosciences ARH1001EA), followed by tyramide signal amplification (TSA) with Opal fluorophores (Akoya Biosciences) diluted 1:100 in 1X plus amplification diluent (Akoya Biosciences FP1498). The primary antibody panel included: CD4 (Mouse monoclonal, clone 4B12, ready-to-use, Agilent, product number IS64930-2), CD8 (Mouse monoclonal, clone C8/144B, ready-to-use, Agilent, product number GA62361-2), FOXP3 (Mouse monoclonal, clone 236A/E7, 1:300, Abcam, product number ab20034), PD1 (Mouse monoclonal, clone

NAT105, ready-to-use, Cell Marque, product number 315M), CD137 (TNFRSF9 or 4-1BB, Mouse monoclonal, clone BBK-2, 1:80, ThermoFisher, product number MA5-13736), cytokeratin (Mouse monoclonal, clone AE1/AE3, ready-to-use, Leica Biosystems, product number NCL-L-AE1/AE). Finally, in the last round, nuclei were counterstained with spectral DAPI (Akoya Biosciences FP1490) and sections mounted with Faramount Aqueous Mounting Medium (Dako S3025).

Each whole-tissue section was scanned on a Vectra-Polaris Automated Quantitative Pathology Imaging System (Perkin Elmer Inc., Waltham, MA, USA). Tissue imaging and spectral unmixing were performed using inForm software (version 2.4.8, Akoya Biosciences), as previously described[54,56].

**Endometrial Carcinoma. Image Analysis using QuPath.**

Image analysis was performed on whole-tissue sections using the open-source digital pathology software QuPath version 0.2.3, as previously described[54]. In short, cell segmentation based on nuclear detection was performed using StarDist 2D algorithm, a method that localizes nuclei via star-convex polygons, incorporated into QuPath software by scripting. A random trees classifier was trained separately for each cell marker by an experienced pathologist annotating the tumor regions. Interactive feedback on cell classification performance, provided during training in the form of markup image, improved the accuracy of machine learning-based phenotyping[2,54]. All phenotyping and subsequent quantifications were performed blinded to the sample identity. Cells close to the border of the images were removed to reduce the risk of artifacts. Based on the fluorescence panels, cells were classified as CD4+ and CD8+ T cells. Subpopulations within CD4+ and CD8+ T cells were then subclassified by the presence and absence of three additional markers: FOXP3, PD1, and CD137. The staining pattern for CD137 considered as positive was described earlier[57]. Cells negative for these markers were defined as "other cell types".

**Implementation details**

Patch contrastive learning is implemented in Python 3.7.3 using Tensorflow 1.14.0[58]. NaroNet is implemented in Python 3.7.3 using PyTorch 1.4.0[59]. Architecture search was performed using ray 1.0.0[46] and hyperpopt 0.2.3.

Synthetic datasets were generated in MATLAB v2019b. Python libraries that were also used include imgaug 0.4.0, tqdm 4.48.2, scipy 1.5.4, numpy 1.18.2, sklearn 0.23.2, seaborn 0.11.0, and pandas 1.1.1. All the experiments were carried out using a server with 16 Intel(R) Xeon(R) E5-2623 v3 @ 3.00GHz CPUs, a RAM of 256 GBs, and 4 GeForce RTX 2080 Ti GPUs of 11GBs. For use as a framework, NaroNet's source code is available on GitHub (https:// https://github.com/djimenezsanchez/NaroNet). For a full description of how to use NaroNet, as well as the most recent versions and dependencies, please refer to the online documentation available on the GitHub page.

**Reporting summary**

Further information on research design is available in the Nature Research Reporting Summary linked to this article.

**Data availability**

All 3 datasets used in this study are publicly available and can be accessed online. All synthetic patient cohorts (including multiplex images, ground-truth masks, and patient data) and high-grade endometrial cancer cohorts (including multiplex images and patient data) are available at Zenodo (https://doi.org/10.5281/zenodo.4596337). Breast cancer cohort is publicly available from the original authors[24] at Zenodo (https://doi.org/10.5281/zenodo.3518284).

**Author contributions**

D.J.S., M.A. and C.O.S. conceptualized the method and planned the experiments. D.J.S. implemented NaroNet, Patch contrastive learning and BioInsights. D.J.S. conducted experimentation on the 3 selected datasets. D.J.S., M.A., C.E.A, and C.O.S. analyzed the results. X.M.G and C.E.A provided patient samples and clinical input throughout the analysis. C.E.A. performed immunohistochemical staining and image acquisition. D.J.S., M.A., and C.O.S conceived the communication and presentation of the method. D.J.S., M.A., C.E.A and C.O.S. wrote the paper with contributions from all coauthors. C.O.S. and H.C. contributed resources and funding acquisition.

**Acknowledgements**

This work was funded by the Spanish Ministry of Science, Innovation and Universities, under grants number RTI2018-094494-B-C22 and RTC-2017-6218-1 (MCIU/AEI/FEDER, UE) (C.O.S.). This work was also funded by the National Cancer Institute (NCI) at the National Institutes of Health (NIH): R01CA184476 (H.C.).